# Cross-Resolution Land Cover Classification Using Outdated Products and Transformers

Huan Ni[a,b,c]*, Yubin Zhao[a,b,c], Haiyan Guan[a,b,c], Cheng Jiang[d], Yongshi Jie[d], Xing Wang[a,b,c], Yiyang Shen[a,b,c]

[a] School of Remote Sensing & Geomatics Engineering, Nanjing University of Information Science & Technology, Nanjing 210044, China
[b] Technology Innovation Center of Integration Applications in Remote Sensing and Navigation, Ministry of Natural Resources, Nanjing 210044, China
[c] Jiangsu Engineering Center for Collaborative Navigation/Positioning and Smart Applications, Nanjing 210044, China
[d] Beijing Key Laboratory of Advanced Optical Remote Sensing Technology, Beijing Institute of Space Mechanics and Electricity, Beijing 100094, China

Large-scale high-resolution land cover classification is a prerequisite for constructing Earth system models and addressing ecological and resource issues. Advancements in satellite sensor technology have led to an improvement in spatial resolution and wider coverage areas. Nevertheless, the lack of high-resolution labeled data is still a challenge, hindering the large-scale application of land cover classification methods. In this paper, we propose a Transformer-based weakly supervised method for cross-resolution land cover classification using outdated data. First, to capture long-range dependencies without missing the fine-grained details of objects, we propose a U-Net-like Transformer based on a reverse difference mechanism (RDM) using dynamic sparse attention. Second, we propose an anti-noise loss calculation (ANLC) module based on optimal transport (OT). Anti-noise loss calculation identifies confident areas (CA) and vague areas (VA) based on the OT matrix, which relieves the impact of noises in outdated land cover products. By introducing a weakly supervised loss with weights and employing unsupervised loss, the RDM-based U-Net-like Transformer was trained. Remote sensing images with 1 m resolution and the corresponding ground-truths of six states in the United States were employed to validate the performance of the proposed method. The experiments utilized outdated land cover products with 30 m resolution from 2013 as training labels, and produced land cover maps with 1 m resolution from 2017. The results show the superiority of the proposed method compared to state-of-the-art methods. The code is available at https://github.com/yu-ni1989/ANLC-Former.

# 1 Introduction

Land surface accounts for 29% of the Earth's total surface area and serves as a major source of carbon emissions and sequestration [1]. Therefore, land cover classification is recognized as a crucial observational variable and is included in the Climate Change Initiative (CCI) program initiated by the European Space Agency [2]. Land cover products not only facilitate the construction of Earth system models for studying past, present, and future climate conditions, but also aids in the identification of distribution patterns in natural landscapes and the exploration of human development activities. With the maturity and development of airborne and spaceborne remote sensing platforms, a large number of high-spatial-resolution remote sensing images are now available for the precise and periodic classification of land cover. However, remote sensing land cover classification still faces the following two challenges:

(1) The influence of human construction activities [3], natural disasters [4], and climate change [5] on the geomorphological environment has caused land cover products to become quickly outdated, necessitating frequent updates.

(2) Despite continuous improvements in the resolution of remote sensing images [6], the corresponding training samples (including images and manual annotations) are difficult to update in a timely manner. Therefore, it is imperative to explore techniques for the effective classification of high-resolution land cover products using low-resolution and outdated products.

In order to overcome the aforementioned issues, Li et al. [7] proposed a low-to-high network for large-scale cross-resolution land cover classification. To prevent the degradation of feature details caused by the dense down-/up-sampling, they employed three different scales of kernels to continuously convolve features while maintaining the input feature size, channels, and resolution. In addition, to meet the cross-resolution mapping objective, they divided the image into confident areas (CA) and vague areas (VA) using confidence propagation (CP) values [8] and optimized the network by computing weakly supervised loss and unsupervised loss for these areas, respectively. However, due to the intrinsic locality of convolutional operations, it is challenging for the network with fixed-scale kernels to learn explicit global and long-range interactions. On the other hand, the CA and VA identification process using CP values does not consider the correspondence between the network prediction and the outdated land cover products. When an area undergoes change due to human activities, the area is not confident according to the outdated land cover products, even though the CP value is high. In this case, the weakly supervised loss is computed using the outdated land cover products, and the optimization process will mislead the network. Therefore, there is room for improvement in the confidence-based CA and VA selection approach for cross-resolution mapping.

In addition, inspired by the tremendous success of Transformers in the field of natural language processing (NLP) [9], researchers have introduced Transformers into the visual field [10]. Dosovitskiy et al. [11] first proposed the Visual Transformer (ViT), which has shown great potential in image recognition tasks due to its exceptional ability to model long-range dependencies. It has also provided a new way to conduct land cover classification using remote sensing images.

Based on the aforementioned developments, in 2017 we proposed a Transformer-based weakly supervised method, aiming to directly classify land covers with a resolution of 1 m using outdated land cover products from 2013 with a resolution of 30 m. In this case, there are two kinds of noises in the outdated land cover products, that is, changes during the period from 2013 to 2017, and misclassifications in the outdated land cover map producing process. To address the problem caused by noises in the outdated products, we proposed an anti-noise loss calculation (ANLC) module. The ANLC module not only considers both aforementioned kinds of noises, but also takes the correspondence issue between network prediction and outdated products into account, which makes the CA and VA selection more reliable. Furthermore, we considered the CP values as weights to improve the weakly supervised loss. To address the issue of missing spatial detail caused by the low resolution of the outdated products, we constructed a U-Net-like Transformer based on the Reverse Difference Mechanism (RDM [12]) which is called RDM-based U-Net-like Transformer for simplicity using dynamic sparse attention. By fusing multiscale features from the encoder through the RDM-based skip connections, the missed spatial details are restored. As a result, the proposed Transformer-based weakly supervised method outperforms existing methods in the specific task.

## 2 Related Works

### 2.1 Classic Land Cover Classification Methods

In the early stages of the development of remote sensing technology, the availability of remote sensing imagery was primarily limited to moderate to low spatial resolutions. Numerous studies have been devoted to the application of machine learning algorithms for low-resolution (LR) land cover classification tasks, including decision trees (DT) [13], support vector machine (SVM) technology [14], and random forests (RF) [15]. These methods utilize the rich spectral information in the image to classify each pixel, generating large-scale low-resolution land cover products.

With the improvement of the spatial resolution of remote sensing images, researchers have gradually shifted their focus to high-resolution land cover classification tasks. Due to limited spectral information and diverse spatial details, traditional pixel-based classifiers have demonstrated poor performance when applied to high-resolution images [16]. These pixel-based methods typically require a plethora of auxiliary data sources, such as elevation, slope, vegetation indices, et cetera, to provide additional information for high-resolution land cover classification tasks [17]. To eliminate the need for additional data, object-based image analysis (OBIA) methods have been proposed. These methods selectively incorporate texture or contextual features obtained from the spatial dimension into the classification process, thereby improving the classification accuracy. However, the manually designed classification scheme hinders the application of OBIA methods in large-scale high-resolution land cover classification [18].

With the rapid development of deep learning technology, convolutional neural networks (CNNs) have made significant progress in pixel-based classification tasks [19], making it possible to generate high-resolution land cover maps accurately and efficiently. However, due to the intrinsic locality of convolutional operations, it is challenging for CNN-based methods to learn explicit global and long-range semantic interactions [20]. Several studies have attempted to address this issue by employing atrous convolutional layers [21], self-attention mechanisms [22], and feature pyramids [23]. However, these methods still have limitations in terms of modeling long-range dependencies.

## 2.2 Land Cover Classification Based on Vision Transformers

Transformers represent a type of neural network architecture that utilizes channel-wise multilayer perceptron (MLP) blocks and attention blocks for cross-location relation modeling. Transformers were initially proposed for NLP [24] and were subsequently introduced into computer vision through pioneering works such as the Detection Transformer (DETR) [25] and the ViT [11]. Compared to CNNs, the most vital difference of Transformers lies in the use of an attention mechanism instead of convolution to achieve global context modeling.

With the rapid development of Transformers in the field of computer vision, many studies have applied Transformers to land cover classification tasks. Gao et al. [26] constructed a systematic framework for multisource remote sensing image processing, which employed typical CNNs for data fusion and then utilized the Spatial-Spectral Vision Transformer (SSViT) for land cover classification, thereby facilitating biodiversity estimation. Yu et al. [27] utilized the rich geometric and spectral characteristics of multispectral light detection and ranging (MS-LiDAR) data and designed a novel Cross-Context Capsule Visual Transformer (CapViT) model for land cover classification. By leveraging three streams of capsule-based transformer encoders, the model learns long-range global feature semantics using different contextual scales to guide land cover type prediction. These methods all achieve accurate land cover type prediction. Ru et al. [28] proposed a weakly supervised semantic segmentation method based on Vision Transformer (ViT) called Token Contrast (ToCo). They devised a Patch Token Contrast module (PTC) to supervise the final patch tokens by utilizing pseudo token relations derived from intermediate layers, which aligned semantic regions to produce more accurate Class Activation Maps (CAM). Subsequently, the Class Token Contrast module (CTC) was employed to further distinguish low-confidence regions in CAM. Finally, the network is trained by utilizing pseudo-labels generated from CAM.

However, traditional attention calculations compute pairwise feature similarity for all spatial positions, resulting in high computational complexity and memory usage, especially for high-resolution inputs [29–31]. Therefore, designing more effective attention mechanisms is an important research direction.

To reduce computation complexity and memory bottlenecks in attention mechanisms, numerous methods have been proposed. In the Swin Transformer, attention is constrained to non-overlapping local windows, and a shifting window operation is introduced to enable information communication between adjacent windows. In order to achieve larger receptive fields within appropriate computational budgets, subsequent works have designed various sparse patterns, such as dilated windows [32] and cross-shaped windows. However, these fixed sparse patterns are handcrafted and lack adaptability to varying data content. By contrast, a BiFormer [33] is a dynamically sparse attention mechanism with query-aware ability, aiming to attend to each query using only a small portion of key-value pairs that are semantically relevant, without scattering attention on irrelevant keys. This effectively reduces computation and memory costs while focusing on the informative parts of the data.

## 2.3 Cross-Resolution Land Cover Classification

The training process of common deep learning methods relies heavily on a large amount of highly accurate annotated data, which is typically generated through laborious and time-consuming manual annotation for specific requirements. Consequently, this severely restricts the mapping coverage range of these methods [34]. By contrast, low-resolution (LR) land cover product data are readily available and abundant, and using LR land cover products as label sources can effectively address this issue. However, due to the resolution mismatch, LR labels may contain a significant number of misclassifications, which can affect the training effectiveness of the network. Therefore, studies have increasingly started to investigate how to map high-resolution land cover in the presence of noise or deficiencies in labels. For example, in cases where only a small amount of HR-labeled data is available, Robinson et al. [35] designed a super-resolution loss function that utilizes 30-meter resolution land cover maps as auxiliary label sources, and produced 1-meter resolution land cover results for the northern Chesapeake region of the United States. However, this approach still requires partial HR labeling during the network training process, and the super-resolution loss is computed based on the statistical relationships between the HR and LR labels, which can only be done when HR labels are available [36].

Li et al. [7] proposed a low-to-high network that employs resolution-preserving backbones to prevent frequent up-/down-sampling, which can degrade feature details. Building upon this, a combination of weakly supervised and unsupervised strategies was utilized to achieve large-scale land cover classification. Chen et al. [37] utilized a graph-based image segmentation algorithm called the Felzenszwalb algorithm to identify homogeneous regions and generate initial seed points based on the spectral features of the image. They then utilized these seed points and superpixels to generate superpixel labels, which were used for optimizing the network and enhancing the resolution of land cover classification. In the Multitemporal Semantic Change Detection track of the 2021 DFC-MSD, due to limited corresponding high-resolution labels, participants were asked to generate land cover changes with a resolution of 1 m using only 30-m resolution land cover labels [38]. Similarly, in the 2022 DFC-SLM, participants were asked to draw land cover maps of many cities in France using only a few labeled data covering very small regions.

In general, achieving large-scale land cover classification using noisy labels with unmatched resolution remains a challenge. Therefore, the exploration of appropriate geospatial data and mapping methods to update and refine large-scale land cover maps with high resolution is of significant importance for the sustainable development of human society [39].

## 3 Study Areas and Data

The data used in this study cover the Chesapeake Bay watershed in the United States, including six administrative states: Maryland (ML), Virginia (VA), Pennsylvania (PA), New York (NY), Delaware (DE), and West Virginia (WV), as illustrated in Figure 1. In more detail, the individual data sources are as follows.

1. The HR (1 m) aerial images from 2017: these images include red (R), green (G), blue (B), and near-infrared (NIR) spectral channels and were obtained from the National Agriculture Imagery Program (NAIP) of the United States Department of Agriculture [40].
2. The LR (30 m) land cover label maps from 2013: these label maps were provided by

the National Land Cover Database (NLCD) of the United States Geological Survey (USGS) [41]. The labels are in the level II classification hierarchy which is shown in the first column in Table 1. These label maps have been re-projected through nearest-neighbor up-sampling to align with the NAIP images.

3. The HR (1 m) ground reference data for the HR (1 m) aerial images from 2017: the data were used for accuracy assessment, and were provided by the Chesapeake Bay Conservancy Land Cover (CCLC) project. The labels in the reference data are shown in the second column in Table 1.

There are 4,000 HR (1 m) aerial images which consist of 3,880×3,880 pixels covering the six states. The HR (1 m) ground reference data were also divided into 4,000 label maps. From Table 1, we can see that the Roads, Buildings and Barren in CCLC category system do not have corresponding categories in the NLCD category system. However, the Roads, Buildings and Barren, as well as the top four categories of the NLCD category system can be considered as Impervious Surface. Therefore, we kept consistent with the existing study [7] which merged the LR land cover labels of 16 classes and HR ground reference of 6 classes into four unified categories, including Tree Canopy (T.C.), Low Vegetation (L.V.), Impervious Surface (I.S.), and Water (W.), as shown in Table 1.

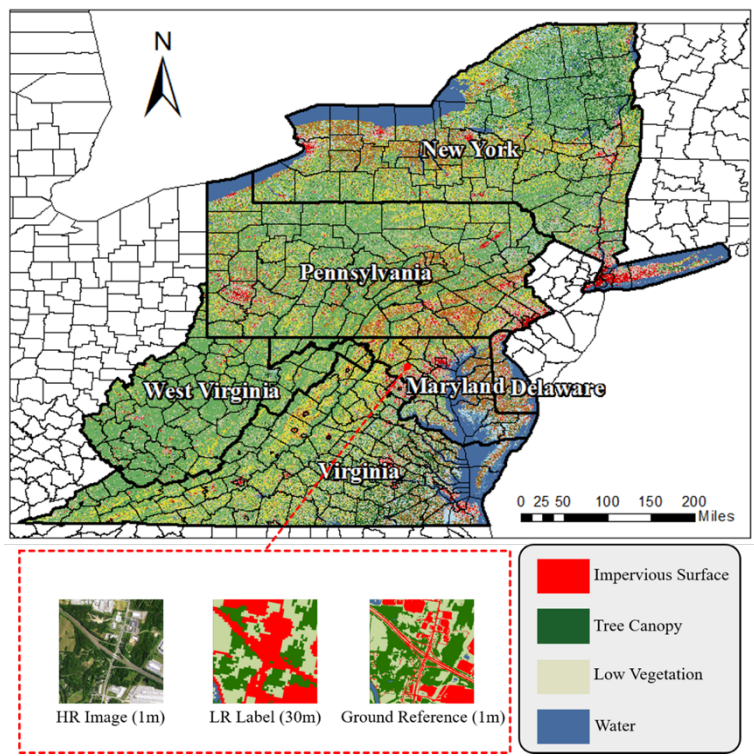

Figure 1 Illustration of the employed datasets [7]

Table 1 Category merging relationship from [7] between land cover data from the National Land Cover Database and ground reference data from the Chesapeake Bay Conservancy Land Cover project.

| NLCD classes | CCLC classes | Target classes |
|---|---|---|
| Developed Open Space | Roads | |
| Developed Low Intensity | Buildings | Impervious Surface |
| Developed Medium Intensity | Barren | |

| | | | | |
|---|---|---|---|---|
| ■ | Developed High Intensity | | | |
| ■ | Deciduous Forest | | | |
| ■ | Evergreen Forest | Tree canopy | ■ | Tree canopy |
| ■ | Mixed Forest | | | |
| ■ | Woody Wetlands | | | |
| ■ | Barren Land | | | |
| ■ | Shrub/Scrub | | | |
| ■ | Grassland/Herbaceous | Low vegetation | ■ | Low vegetation |
| ■ | Pasture/Har | | | |
| ■ | Cultivated Crops | | | |
| ■ | Emergent Herbaceous Wetlands | | | |
| ■ | Open Water | Water | ■ | Water |

## 4 Method

This paper proposes a Transformer-based weakly supervised method that leverages outdated LR land cover products with noises to achieve large-scale HR land cover maps. As illustrated in Figure 2, the entire workflow comprises two modules: an RDM-based U-Net-like Transformer and ANLC. In this section, we sequentially introduce the two modules.

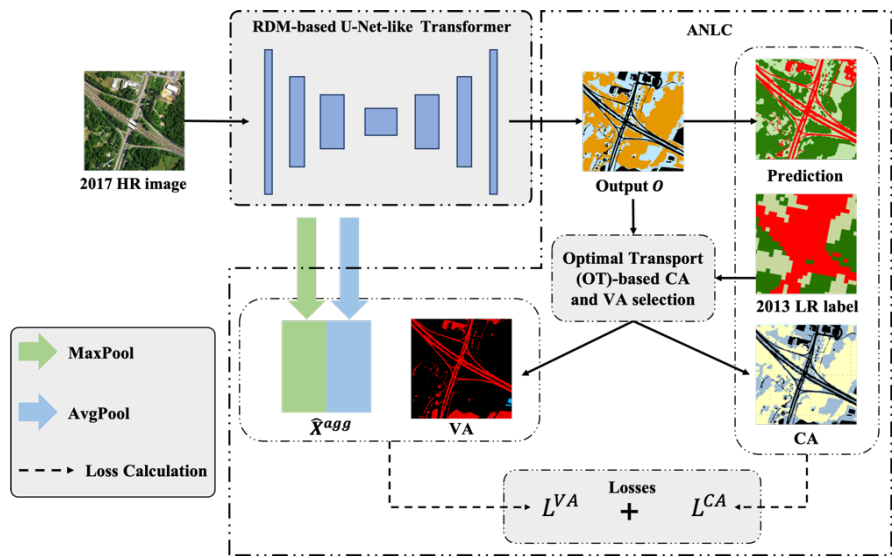

Figure 2. Overall workflow of the proposed framework

### 4.1 An RDM-based U-Net-like Transformer

#### 4.1.1 Overall Architecture

The overall architecture is shown in Figure 3. The network consists of an encoder, a bottleneck, a decoder, and RDM-based skip connections. The basic unit of the network is the BiFormer block [33]. Transformer units have a great capacity for long-range dependency modeling. However, the features extracted by Transformer units lack spatial details. To address this issue, we proposed the RDM[12]-based skip connection to aid Transformer units. Not only can RDM enhance the spatial details, but it can also relieve the predominant effects of global information. In this way, the proposed RDM-based U-Net-like Transformer takes both the global information and the fine-grained details into consideration, with the help of the RDM-based skip connection. During the encoding stage, the input image $X \in \mathcal{R}^{W \times H \times 4}$ is transformed into sequence embeddings $X' \in \mathcal{R}^{\frac{W}{4} \times \frac{H}{4} \times 48}$, using overlapping patch embeddings [30]. Then, a linear embedding layer is applied to project the feature into $\mathcal{R}^{\frac{W}{4} \times \frac{H}{4} \times C}$ space and obtain the first-level feature, $X_1^e \in \mathcal{R}^{\frac{W}{4} \times \frac{H}{4} \times C}$, through a BiFormer block. Next, using $X_1^e$ as input, a patch-merging layer, which down-sample inputs and the second BiFormer block, are performed to generate the second-level feature, $X_2^e \in \mathcal{R}^{\frac{W}{8} \times \frac{H}{8} \times 2C}$. After repeating the operations, including the patch-merging layer and the BiFormer block, the third-level feature, $X_3^e \in \mathcal{R}^{\frac{W}{16} \times \frac{H}{16} \times 4C}$ is obtained. Then, $X_3^e$ is put into the bottleneck which is composed of two BiFormer blocks so that the high-level feature $X^h \in \mathcal{R}^{\frac{W}{32} \times \frac{H}{32} \times 8C}$ is generated.

Inspired by the U-Net network [42], we designed a symmetric Transformer-based decoder. The decoder consists of several BiFormer blocks and patch-expanding layers. In contrast to the patch-merging layer, a patch-expanding layer is specially designed to perform up-sampling. Similar to the Encoder, the Decoder generates $\hat{X}_3^d \in \mathcal{R}^{\frac{W}{16} \times \frac{H}{16} \times 4C}$, $\hat{X}_2^d \in \mathcal{R}^{\frac{W}{8} \times \frac{H}{8} \times 2C}$, and $\hat{X}_1^d \in \mathcal{R}^{\frac{W}{4} \times \frac{H}{4} \times C}$, in sequence. The difference is that the RDM-based skip connection enhanced the initial features, $X_3^d \in \mathcal{R}^{\frac{W}{16} \times \frac{H}{16} \times 4C}$, $X_2^d \in \mathcal{R}^{\frac{W}{8} \times \frac{H}{8} \times 2C}$, and $X_1^d \in \mathcal{R}^{\frac{W}{4} \times \frac{H}{4} \times C}$, generated by the inner layers of the Decoder. The RDM-based skip connection takes $X^h$ which is generated by the bottleneck, and a lower-level feature ($X_2$, $X_2$, or $X_3$) as inputs, and generates the spatial details $X_i^{sd}$ to assist the BiFormer. Taking the $X_3^{sd}$ as an example, $X_3^{sd}$ is then fused with $X_3^d$ so that $\hat{X}_3^d$ is produced. Finally, the last patch-expanding layer is used to perform 4 ×up-sampling to restore the resolution of the feature maps to the input resolution, and then a linear projection layer is applied to generate the pixel-wise segmentation output $O \in \mathcal{R}^{W \times H \times K}$, where $K$ is the number of categories. Below we describe the key components including BiFormer and RDM-based skip connection in detail.

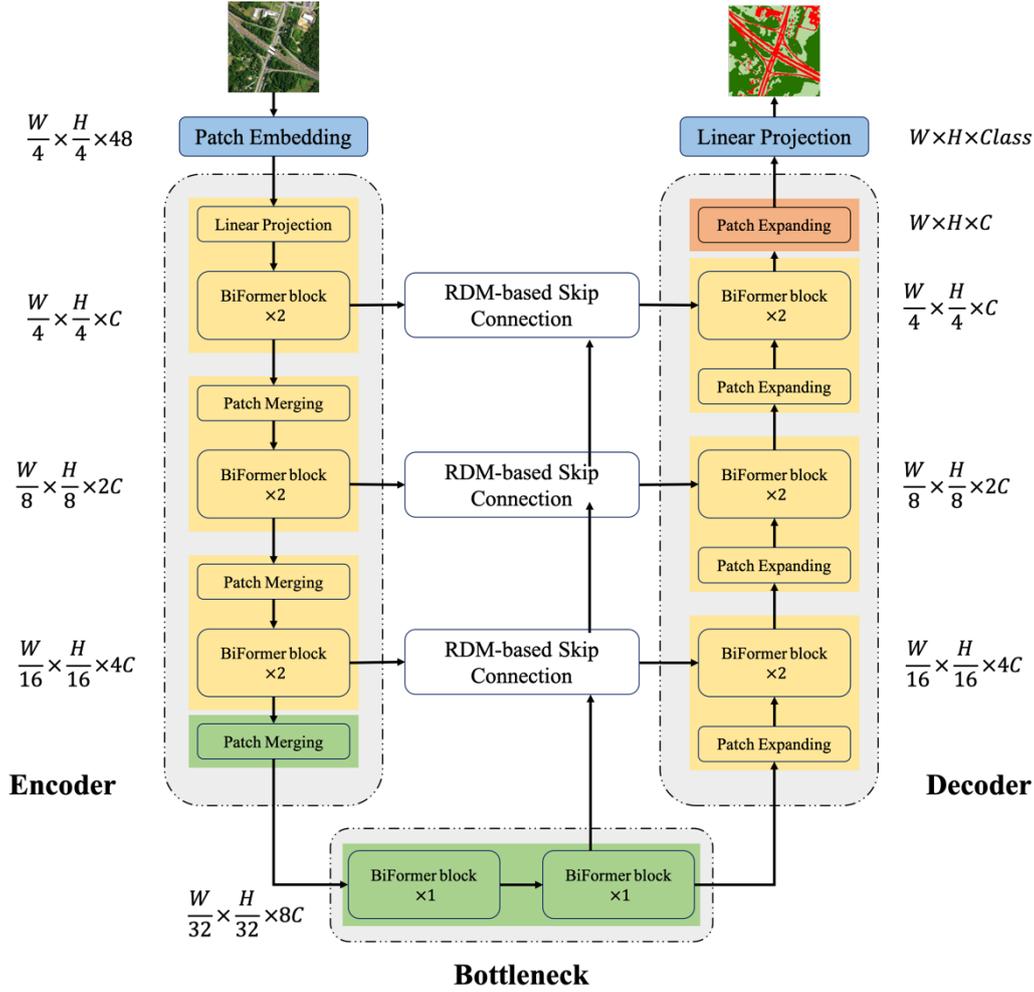

Figure 3 The overall architecture of Unet-like Transformer

### 4.1.2 BiFormer Block

To alleviate the high computational complexity and heavy memory cost issues of multi-head self-attention models (MHSA), different sparse attention mechanisms are proposed. In this study, we employed the BiFormer [33] to meet this objective. The BiFormer is constructed based on a dynamic and query-aware sparse attention mechanism. The key idea is to filter out the most irrelevant key-value pairs in the coarse region so that only a small portion of confident regions are kept. Then, fine-grained token-to-token attention in the union of these confident regions is constructed. Figure 4 illustrates the details of the BiFormer Block.

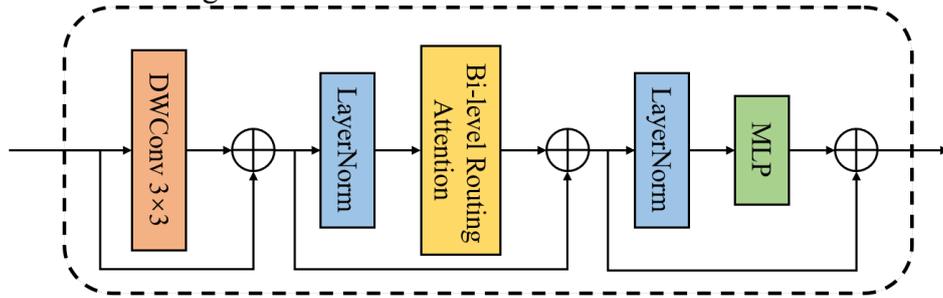

Figure 4. The BiFormer block

First, a 3 × 3 depth-wise convolution DWConv is utilized to encode relative positions implicitly. Subsequently, a Bi-level Routing Attention module and a two-layer MLP with GELU non-linear activations are applied for cross-location relation modeling and per-location embedding, respectively. The detailed explanation of the Bi-level Routing Attention as follows.

Given a two-dimensional (2D) input feature map, $X \in \mathcal{R}^{H \times W \times C}$, it is first divided into $S \times S$ non-overlapping regions, each containing $\frac{HW}{S^2}$ feature vectors. At this time, $X$ is reshaped into $X^s \in \mathcal{R}^{S^2 \times \frac{HW}{S^2} \times C}$. Subsequently, the query ($Q$), key ($K$), value ($V$) tensor, $Q, K, V \in \mathcal{R}^{S^2 \times \frac{HW}{S^2} \times C}$, are obtained. Then, a directed graph is constructed to determine which regions should attend to each query region. Specifically, the average values of $Q$ and $K$ for each region are calculated to obtain the region-level queries and keys $Q^r$ and $K^r$. Then, the adjacency matrix $A^r \in \mathcal{R}^{S^2 \times S^2}$ of the region-to-region affinity graph is obtained through the matrix multiplication of $Q^r$ and the transpose of $K^r$:

$$A^r = Q^r \cdot (K^r)^T. \tag{1}$$

The elements in the adjacency matrix represent the degree of association between the two regions. Based on the adjacency matrix, the top $k$ connections with the highest degree of association are selected for each region. Specifically, we obtain an index matrix, $I^r \in N^{S^2 \times k}$, with the row-wise top $k$ operator:

$$I^r = \text{argwhere}\left(\text{Top}^k(A^r)\right). \tag{2}$$

Therefore, the $i^{th}$ row of the matrix $I^r$ contains the indices of the top $k$ most relevant regions for the $i^{th}$ region.

With the region-to-region index matrix $I^r$, key and value tensors ($K^g, V^g \in \mathcal{R}^{S^2 \times \frac{kHW}{S^2} \times C}$) can be collected for computing fine-grained token-to-token attention:

$$X = Attention(Q, K^g, V^g) + LCE(V). \tag{3}$$

Here, $LCE(\cdot)$ is a local context enhancement function parameterized by deep convolutions with a kernel size of 5. The specific calculation is as follows:

$$LCE(V) = Conv_{5 \times 5}(V). \tag{4}$$

The stride of the convolution operation is set to 1, padding is 2, and $V$ is the value tensor.

### 4.1.3 RDM-based Skip Connection

Skip connections can relieve the impact of the ignorance of spatial details caused by down-sampling. However, the spatial details are still pushed aside due to the predominant effect of global information. To ensure the equal treatment of global information and spatial details when the U-Net-like Transformer is backward, we integrated RDM [12] into the skip connection to meet this objective. Suppose the RDM-based skip connection is performed in the $i$-th layer, $1 \leq i \leq 3$, as shown in Figure 3, the RDM containing cosine and neural alignments transforms the high-level feature map, $X^h \in \mathcal{R}^{\frac{W}{32} \times \frac{H}{32} \times 8C}$, according to the $i$-th level feature map

$X_i^e \in \mathcal{R}^{W_i^e \times H_i^e \times C_i^e}$ in the $i$-th layer of the Encoder, resulting in $X_i^{h,cos}$ and $X_i^{h,neu}$. The cosine alignment has a strict mathematical explanation. It uses the cosine function to compute a range of weights for the channels of $X^h$. Large weights are given to the channels of $X^h$ when these channels of features are more similar to a certain channel of $X_i^e$. Then, a matrix multiplication can transform $X^h$ as $X_i^{h,cos}$, which is aligned to $X_i^e$. The neural alignment meets a similar objective using convolutions. The detailed process can be found in [12].

Subsequently, a difference map which contains the spatial details is obtained through the differential operator, as shown in formulas (5) and (6). Finally, concatenation is performed, followed by a ReLU function to set all negative values to 0, thus excluding global information and highlighting the information of spatial details. The specific process is as follows:

$$X_i^{diff,cos} = S(X_i^e) - S(X_i^{h,cos}), \tag{5}$$

$$X_i^{diff,neu} = S(X_i^e) - S(X_i^{h,neu}), \tag{6}$$

$$X_i^{sd} = \text{ReLU}\left(\text{Cat}(X_i^{diff,cos}, X_i^{diff,neu})\right), \tag{7}$$

where $S(\cdot)$ is the Sigmoid function, which transforms the feature values into a range from 0 to 1, and $\text{Cat}(\cdot)$ is the feature concatenation operation.

Finally, $X_i^{sd}$ is connected to the $i$-th level feature map $X_i^d$ in the Decoder. In this way, the spatial details are recovered, and an enhanced feature map of the $i$-th layer in the Decoder is generated:

$$\hat{X}_i^d = \text{Linear}\left(\text{Cat}(X_i^{sd}, X_i^d)\right), \tag{8}$$

where Cat $(\cdot,\cdot)$ concatenates $X_i^{sd}$ and $X_i^d$, and Linear $(\cdot)$ is a linear layer which compresses the concatenated features.

## 4.2 Anti-Noise Loss Calculation

The disparity in both the time and resolution between HR ground-truths in 2017 and outdated LR labels in 2013 leads to a significant amount of noise in the LR data, which in turn adversely affects the training effectiveness of the model. Therefore, during this stage, we proposed an ANLC module to produce more reliable losses for back-propagation.

Anti-noise loss calculation divides the output $O \in \mathcal{R}^{W \times H \times K}$ of the RDM-based U-Net-like Transformer into confident areas (CA) and vague areas (VA) according to the predicted confidence probability (CP) values. Unlike the method which sets a fixed threshold to make this division in the study [7], we proposed an OT-based CA and VA selection method. The main drawback of the method in [7] is that it only takes the output of the network into consideration. Because outdated land cover products and current satellite imagery are not on the same time, some areas may have changed over time. Even if the network prediction $P$ generated by $O$ are accurate, when calculating losses using outdated land cover products, the predictions for these areas are still considered wrong, misleading the network training process. Therefore, the CA and VA selection should take both the predictions and the coarse labels provided by the outdated land cover product into consideration. However, we still cannot know which areas changed over time. In this case, we introduced OT to alleviate the impact of this problem as much as possible from the perspective of global optimization.

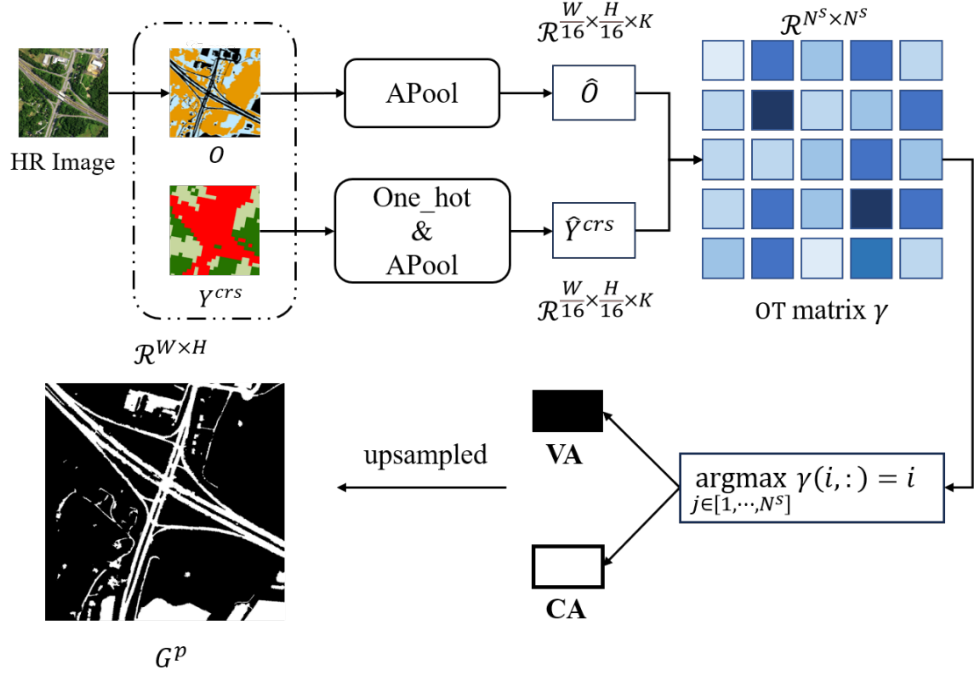

Figure 5. The flowchart of the anti-noise loss calculation.

First, the OT-based CA and VA selection transforms the coarse label map $Y^{crs}$ provided by the outdated land cover product into one-hot form $Y^{crs} \in \mathcal{R}^{W \times H \times K}$. Subsequently, we averagely pooled the $Y^{crs}$ and $O$ into the $\mathcal{R}^{\frac{W}{16} \times \frac{H}{16} \times K}$ space simultaneously as follows:

$$\hat{Y}^{crs} = \text{APool}(Y^{crs}), \tag{9}$$
$$\hat{O} = \text{APool}(O). \tag{10}$$

This average pooling operator gains category interactions for $\hat{Y}^{crs}$ and $\hat{O}$. Then, $\hat{Y}^{crs}$ and $\hat{O}$ were transformed into a set of vectors in the form of $\mathcal{R}^{N^s \times K}$, where $N^s = \frac{W}{16} \times \frac{H}{16}$. The algorithm from [43] is utilized to obtain the OT matrix $\gamma \in \mathcal{R}^{N^s \times N^s}$ between $\hat{Y}^{crs}$ and $\hat{O}$. If the $P$ and $Y^{crs}$ are both correct, the largest value in each row or column in $\gamma$ should be on the diagonal. Otherwise, either the prediction is wrong or the coarse label is wrong. In this way, to determine the CA and VA, a binary mask $G^v \in \mathcal{R}^{N^s \times 1}$ was obtained as follows:

$$\begin{cases} G^v(i) = 1, & \text{if } \underset{j \in [1, \cdots, N^s]}{\arg\max} \gamma(i,:) = i; \\ G^v(i) = 0, & \text{otherwise}. \end{cases} \tag{11}$$

Then, $G^v$ was transformed into the image plane $G^p \in \mathcal{R}^{\frac{W}{16} \times \frac{H}{16}}$. Finally, $G^p$ was up-sampled into the original image space $\mathcal{R}^{W \times H}$ using nearest-neighbor sampling so that the CA and VA were finally determined:

$$\begin{cases} (i,j) \text{ belongs to CA}, & \text{if } G^p(i,j) = 1; \\ (i,j) \text{ belongs to VA}, & \text{otherwise}. \end{cases} \tag{12}$$

For the CA, the weakly supervised loss $L^w$ with weights was computed as follows:

$$L^{CA} = \frac{1}{H \times W} \left( \sum_{i=1}^{H} \sum_{j=1}^{W} G^p(i,j) \times W^c(i,j) \times \left( -O(i,j)[Y^{crs}(i,j)] + \log\left( \sum_{k=1}^{K} e^{O(i,j)[k]} \right) \right) \right), \tag{13}$$

where $W^c(i,j)$ is the weight which is set as the confidence that the pixel on position $(i,j)$ belongs to a certain category:

$$W^c(i,j) = \exp\left( \max_{k \in [1,\cdots,K]} \text{Softmax}(O(i,j,:)) \right), \tag{14}$$

where $\exp(\cdot)$ computes the exponent and $\text{Softmax}(\cdot)$ is the Softmax transformation. For the VA, the unsupervised DVA loss $L^{VA}$ [7] was employed. $L^{VA}$ was used to measure the inter-area variance between CA and VA within a category. Before the computation of $L^{VA}$, $\hat{X}_1^d$ was aggregated using the maximum and average pooling operators ($\text{MPool}(\cdot)$ and $\text{APool}(\cdot)$) as follows:

$$\begin{cases} \hat{X}^{max} = \text{MPool}(\hat{X}_1^d); \\ \hat{X}^{avg} = \text{APool}(\hat{X}_1^d); \\ \hat{X}^{agg} = \text{Cat}(\hat{X}^{max}, \hat{X}^{avg}). \end{cases} \tag{15}$$

where $\text{Cat}(\cdot,\cdot)$ concatenates $\hat{X}^{max}$ and $\hat{X}^{avg}$. For category $k$, the mean vector $\bar{X}_k^{agg}$ of all the features of $\hat{X}^{agg}$, belonging to category $k$, the mean vector $\bar{X}_k^{agg,ca}$ of all features of $\hat{X}^{agg}$ in CA belonging to category $k$, and the mean vector $\bar{X}_k^{agg,va}$ of all features of $\hat{X}^{agg}$ in VA belonging to category $k$ were computed. Then, the variance within category $k$ was computed as follows:

$$\sigma_k^2 = p^{ca} \|\bar{X}_k^{agg,ca} - \bar{X}_k^{agg}\|_2^2 + p^{va} \|\bar{X}_k^{agg,va} - \bar{X}_k^{agg}\|_2^2, \tag{14}$$

where $p^{ca}$ and $p^{va}$ represent the proportions of the CA and VA to the whole area in the image plane, respectively. The $\|\cdot\|_2^2$ computes the 2-norm. Finally, $L^{VA}$ was computed as follows:

$$L^{VA} = \sum_{k=1}^{K} \sigma_k^2. \tag{15}$$

Combining $L^{CA}$ and $L^{VA}$, the anti-noise loss $L^{AnN}$ using the OT-based CA and VA selection was obtained as follows:

$$L^{AnN} = L^{CA} + L^{VA}. \tag{16}$$

By iteratively minimizing $L^{AnN}$, the proposed RDM-based U-Net-like Transformer was optimized. Using the trained RDM-based U-Net-like Transformer, up-to-date land cover maps were obtained.

## 5 Experiments

The experiments were conducted on six datasets, as described in section 3. Each dataset was obtained in a state of the Chesapeake Bay watershed. Experiments were performed on these datasets. The results of the land cover classification were visually and quantitatively analyzed using ground references to evaluate the effectiveness of the proposed method, and allow for comparison with existing methods.

## 5.1 Experimental Settings

The proposed method was implemented using Python 3.8 and PyTorch 1.12.1. The Adam Weight Decay Optimizer (AdamW) was utilized, with an initial learning rate of 0.01. If the loss showed no further improvement over a span of eight training epochs, the initial learning rate was reduced to 10% of the current learning rate. During the training process, each tile was randomly cropped to 100 patches with a size of $512 \times 512$. The model was trained for 50 epochs with a batch size of four. To evaluate the performance of the proposed method, we employed quantitative metrics calculated between the HR ground reference and the predicted results. These metrics included the mean Intersection over Union (mIoU), the Frequency Weighted Intersection over Union (FWIoU), the Kappa coefficient, and the Overall Accuracy (OA).

In order to validate the superiority of the proposed method, we compared the proposed method with four different existing weakly supervised methods, namely DeepLabv3+ [44], HRNet [45], ToCo [28], and L2HNet [7]. Specifically, L2HNet is composed of a resolution-preserving (RP) backbone, a confident area selection (CAS), and a low-to-high (L2H) loss $L_{L2H}$.

## 5.2 Results and Comparisons

First, to provide a detailed comparison of the visual results obtained from different methods, the results for selected regions within the six states of the Chesapeake Bay watershed are presented in figure 6. We specifically chose representative results that cover complex scenes in these states. For quantitative comparisons, all accuracy metrics were computed based on HR ground references. During the training process, only LR labels were used for supervision. As a result, the final evaluation results reflect the model's performance in learning the accurate HR representation using only HR images and LR labels. Tables 2–7 present the quantitative comparisons in the six states.

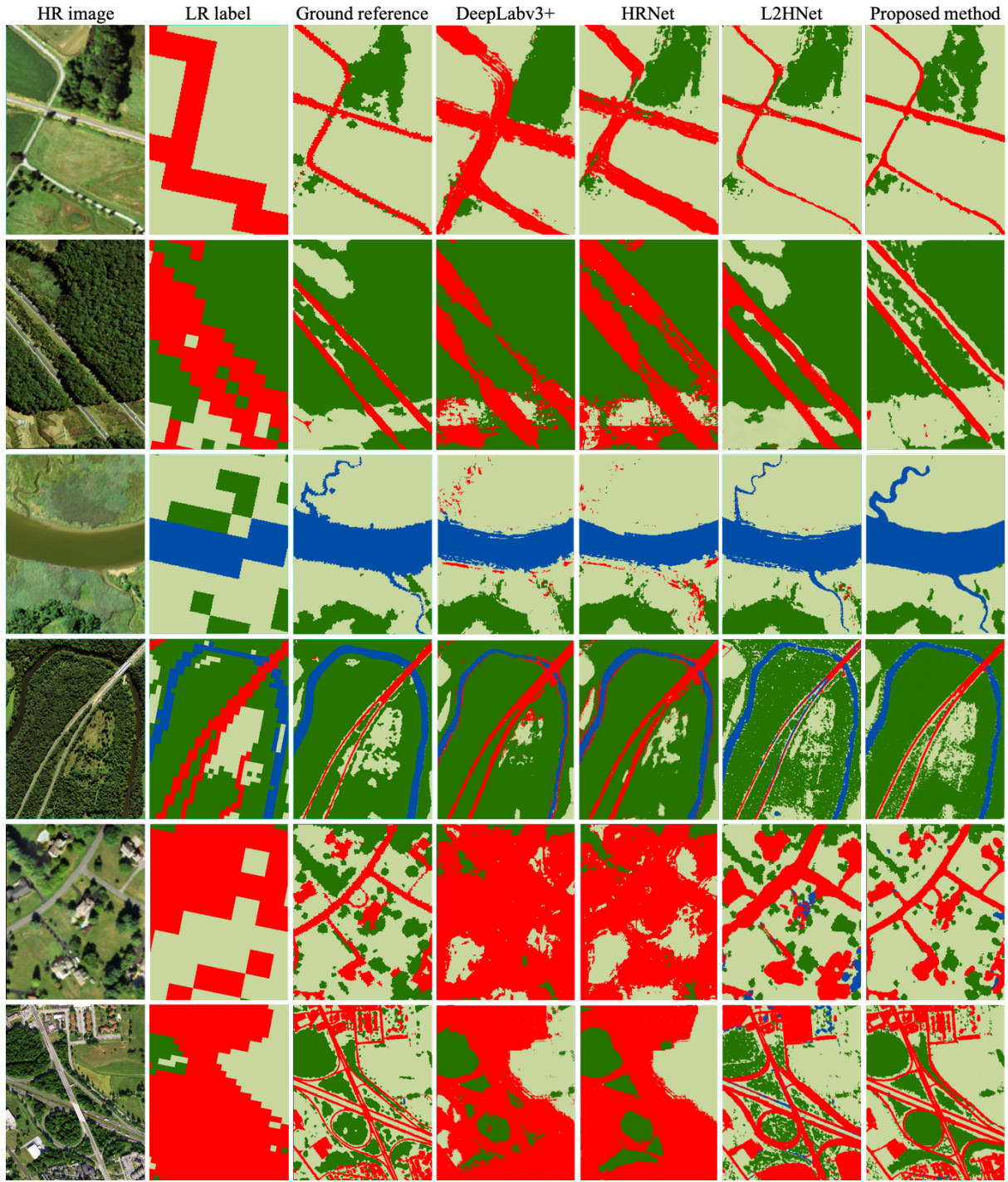

Figure 6. The visual comparison between different methods using selected regions within the six states

**5.2.1 Visual Comparisons**

The fourth and fifth columns in figure 6 illustrate that the predicted results of DeepLabv3+ and HRNet are close to the LR labels, although some rivers are wrongly predicted as impermeable surfaces. There are primarily two reasons for this observation. First, these models do not have a denoising module to relieve the noisy issue of the coarse labels provided by outdated land cover products. As a result, they consider the coarse labels as strong supervision and directly compute the cross-entropy (CE) loss for the entire region. Second, the repeated

up-/down-sampling operations in the models result in the deterioration of the fine-grained details of the features, leading to a significant loss of spatial information. This validates that obtaining sufficient land cover classification results using outdated land cover products requires specific operations for the weak supervision.

As shown in the sixth column in figure 6, L2HNet has improved the mapping results to some extent compared to DeepLabv3+ and HRNet. L2HNet is designed using the weakly supervised strategies to deal with the noise issue of coarse labels. This validates the positive influence of these weakly supervised strategies.

Specifically, L2HNet demonstrates really good land cover classification results compared to traditional deep learning methods, with predicted contours closer to the HR imagery. However, due to the limitations of convolutional operations, this approach fails to achieve global modeling, resulting in significant false positives. For example, shadows are incorrectly classified as impermeable surfaces or water. In addition, the instability in confidence region selection leads to problems such as insufficiently smooth contour edges.

The last column in figure 6 is the classification results of our method. Our method shows obvious advantages in terms of visual comparison. First, by utilizing RDM-based skip connections, the network not only extracts global information but also recovers spatial information loss caused by down-sampling. As a result, the mapped output maintains the same resolution as the input image, encompassing more details of land cover. The contours of roads and rivers are also smoother and more complete. Second, with the improved influence of the confidence region selection module, the predicted results are less affected by noise in coarse labels, approaching the HR imagery and even providing richer details than L2HNet. Comparing the results across the six selected regions, the proposed method demonstrates an accurate and robust land cover classification capability across different scenarios, while preserving a greater level of detail in land cover. Furthermore, addressing the resolution gap, the proposed method not only alleviates the overfitting issues presented in traditional networks (i.e., DeepLabv3+ and HRNet), but also reduces inconsistent noise in the predicted results through the use of the OT-based confidence region selection module. Compared to L2HNet, our method yields more accurate edge results and fewer errors, demonstrating its superiority in cross-resolution land cover classification tasks using outdated products.

Summarizing the visual comparison, the traditional deep learning methods (i.e., DeepLabv3+ and HRNet) employed for comparison show inferior land cover classification performance, with blurry edges and significant loss of detail when directly utilizing coarse labels as supervision sources. At the same time, although L2Hnet is capable of generating land cover classification results with precise edge delineation, it still exhibits weaker performance in the scenes with sparsely distributed and confusing land cover categories, resulting in erroneous predictions. The proposed Transformer-based weakly supervised method, however, exhibits excellent performance, yielding mapping results with accurate land cover classification and precise delineation of land objects.

### 5.2.2 Quantitative Comparisons

For quantitative comparisons, all accuracy metrics were computed based on HR ground references. During the training process, only LR labels were used as the sole supervision source. As a result, the final evaluation results reflect the model's performance in learning the accurate

HR representation based on inputs of HR images and LR labels. Tables 2–7 present the quantitative comparison results of the six states. To ensure a comprehensive comparison, we present the mIoU, FWIoU, Kappa, and OA of all methods on the datasets of the six states in tables 2–7 as well as the IoU score of each category. For ease of comparison, we calculate the average accuracy values across the six states in table 8.

Due to the specific nature of cross-resolution land classification tasks, traditional encoder-decoder architectures or deep convolutional frameworks such as DeepLabv3+ and HRNet suffer from loss of spatial information during repeated up-/down-sampling operations, leading to deterioration in fine-grained detail prediction. Moreover, employing LR labels directly as strong supervision sources encourages prediction results to align with coarse labels. As a result, these approaches fail to obtain accurate predictions. A considerable number of label errors are present in the results shown in figure 6. Consequently, the IoU scores on impervious surfaces and water are really low. Higher scores are observed for tree canopy and low vegetation. Due to the imbalanced performance across different categories, DeepLabv3+ and HRNet achieved lower mIoU, Kappa, and FWIoU scores compared with other methods.

Compared to traditional deep learning network approaches, the latest cross-resolution methods show improved performance. L2Hnet demonstrates better results in the categories of impervious surfaces and water, achieving IoU scores of approximately 0.39 and 0.65, respectively, across the datasets from the six states. This is attributed to its strategy of combining a weakly supervised learning strategy and a resolution preserving backbone, which help mitigate the impact of the noise in coarse labels. However, due to temporal mismatch, instability is present in the selection of confidence areas, which means that even if the prediction is correct, it can be misled by erroneous labels. Hence, L2HNet does not exhibit competitive scores for tree canopy and low vegetation, with IoU scores of approximately 0.75 and 0.65, respectively. It also fails to consistently demonstrate good performance across all datasets, leading to fluctuations in IoU scores between different experimental regions. For example, the IoU score for low vegetation is only 0.69 in the Delaware dataset, whereas it is as high as 0.81 in Virginia.

Compared to other methods, the proposed Transformer-based weakly supervised method tested in this study demonstrates balanced performance, achieving the highest IoU score for each category. Specifically, the average IoU scores for impervious surfaces, low vegetation, tree canopy, and water are 0.47, 0.73, 0.81, and 0.70, respectively. These scores significantly outperform traditional deep learning methods and demonstrate superiority compared to cross-resolution mapping approaches. In terms of class proportion-insensitive metrics, namely mIoU and OA, the average values are approximately 0.68 and 0.85, respectively, which are significantly better than those of traditional models. For class proportion-sensitive metrics, namely FWIoU and Kappa, the average values are approximately 0.76 and 0.73, respectively.

It is evident that the method proposed in this paper achieves a balanced performance across all categories. This is attributed to the RDM-based U-Net-like Transformer backbone, which effectively retains more information about spatial details, thereby improving the performance on the edges of each category. In addition, ANLC effectively reduces the impact of noise in outdated land cover products by incorporating confidence regions determined by the OT matrix. Therefore, the method presented in this paper is well-suited to the task of cross-resolution land cover classification utilizing outdated samples.

To further investigate the performance of the proposed RDM-based U-Net-like Transformer backbone and ANLC, we combined the resolution-preserving (RP) backbone [7] of L2HNet with ANLC (RP+ANLC), and combined the RDM-based U-Net-like Transformer ($UNet_{RDM+T}$) backbone with CAS and L2H loss ($UNet_{RDM+T}$+CAS+$L_{L2H}$). By comparing L2HNet (RP+CAS+$L_{L2H}$) with RP+ANLC, it can be observed that RP+ANLC achieves greater accuracy than L2HNet in terms of the results achieved across six states. This demonstrates that ANLC is more powerful than CAS+$L_{L2H}$. Anti-noise loss calculation considers both network predictions and coarse labels, which helps filter out errors in outdated samples. As a result, the IoU for each class is increased by 0.3 or even more. On the other hand, when comparing the $UNet_{RDM+T}$+CAS+$L_{L2H}$ with the proposed method ($UNet_{RDM+T}$+ANLC), it is apparent that the performance of the network decreases after relacing ANLC using the CAS+L2H loss. This further confirms the effectiveness of ANLC in improving the overall performance of the network. Comparing the proposed method with RP+ANLC, it is clear that the RDM-based U-Net-like Transformer backbone is more effective in feature representation compared to the RP backbone of L2HNet. The improved performance of the proposed method can be attributed to the ability of the RDM-based skip connection to capture more detailed information, leading to enhanced feature representation and overall classification accuracy.

Summarizing the quantitative analysis, traditional deep learning methods fail to meet the requirements for large-scale cross-resolution land cover classification. The cross-resolution model L2HNet has demonstrated superior performance. However, the localized nature of convolution operations and the information loss caused by down-sampling have resulted in varying degrees of misclassification and loss of detail. On the other hand, the performance of L2Hnet is further constrained by the noise presented in outdated land cover products. The proposed method, by comparison, demonstrates more balanced scores and outperforms other comparison methods. Overall, both visual comparison and quantitative comparison of results indicate the superiority of the proposed method in large-scale cross-resolution land cover classification tasks utilizing outdated land cover products.

Table 2 Quantitative comparisons in Maryland (ML)

| State | Method | mIoU | FWIoU | Kappa | OA | Class | | | |
|---|---|---|---|---|---|---|---|---|---|
| | | | | | | I.S. | L.V. | T.C. | W. |
| ML | DeepLabv3+ | 0.52 | 0.62 | 0.61 | 0.64 | 0.25 | 0.58 | 0.72 | 0.55 |
| | HRNet | 0.53 | 0.63 | 0.63 | 0.75 | 0.24 | 0.62 | 0.67 | 0.58 |
| | L2HNet (RP+CAS+$L_{L2H}$) | 0.62 | 0.71 | 0.65 | 0.82 | 0.36 | 0.72 | 0.79 | 0.60 |
| | RP+ANLC | 0.65 | 0.73 | 0.72 | 0.83 | 0.42 | 0.70 | 0.81 | 0.66 |
| | $UNet_{RDM+T}$+CAS+$L_{L2H}$ | 0.66 | 0.73 | 0.72 | 0.84 | 0.45 | 0.71 | 0.82 | 0.65 |
| | Proposed | 0.70 | 0.76 | 0.75 | 0.88 | 0.55 | 0.79 | 0.83 | 0.64 |

Table 3 Quantitative comparisons in Pennsylvania (PA)

| State | Method | mIoU | FWIoU | Kappa | OA | Class | | | |
|---|---|---|---|---|---|---|---|---|---|
| | | | | | | I.S. | L.V. | T.C. | W. |
| PA | DeepLabv3+ | 0.55 | 0.68 | 0.65 | 0.79 | 0.28 | 0.60 | 0.74 | 0.58 |
| | HRNet | 0.52 | 0.59 | 0.58 | 0.72 | 0.23 | 0.61 | 0.66 | 0.60 |

| | L2HNet (RP+CAS+$L_{L2H}$) | 0.61 | 0.70 | 0.72 | 0.81 | 0.39 | 0.66 | 0.75 | 0.66 |
| | RP+ANLC | 0.64 | 0.71 | 0.71 | 0.82 | 0.40 | 0.68 | 0.80 | 0.68 |
| | $UNet_{RDM+T}$+CAS+$L_{L2H}$ | 0.63 | 0.72 | 0.73 | 0.82 | 0.42 | 0.68 | 0.78 | 0.65 |
| | Proposed | 0.66 | 0.74 | 0.73 | 0.84 | 0.45 | 0.70 | 0.80 | 0.67 |

Table 4 Quantitative comparisons in Virginia (VA)

| State | Method | mIoU | FWIoU | Kappa | OA | Class | | | |
|---|---|---|---|---|---|---|---|---|---|
| | | | | | | I.S. | L.V. | T.C. | W. |
| VA | DeepLabv3+ | 0.52 | 0.66 | 0.61 | 0.75 | 0.24 | 0.60 | 0.77 | 0.45 |
| | HRNet | 0.56 | 0.68 | 0.65 | 0.79 | 0.27 | 0.62 | 0.78 | 0.58 |
| | L2HNet (RP+CAS+$L_{L2H}$) | 0.61 | 0.60 | 0.68 | 0.82 | 0.37 | 0.66 | 0.81 | 0.61 |
| | RP+ANLC | 0.63 | 0.73 | 0.71 | 0.84 | 0.40 | 0.67 | 0.80 | 0.64 |
| | $UNet_{RDM+T}$+CAS+$L_{L2H}$ | 0.65 | 0.75 | 0.70 | 0.85 | 0.38 | 0.70 | 0.84 | 0.67 |
| | Proposed | 0.68 | 0.76 | 0.72 | 0.87 | 0.44 | 0.74 | 0.86 | 0.68 |

Table 5 Quantitative comparisons in New York (NY)

| State | Method | mIoU | FWIoU | Kappa | OA | Class | | | |
|---|---|---|---|---|---|---|---|---|---|
| | | | | | | I.S. | L.V. | T.C. | W. |
| NY | DeepLabv3+ | 0.49 | 0.50 | 0.50 | 0.68 | 0.26 | 0.51 | 0.64 | 0.53 |
| | HRNet | 0.52 | 0.58 | 0.58 | 0.73 | 0.30 | 0.56 | 0.62 | 0.58 |
| | L2HNet (RP+CAS+$L_{L2H}$) | 0.60 | 0.68 | 0.63 | 0.78 | 0.40 | 0.59 | 0.72 | 0.72 |
| | RP+ANLC | 0.64 | 0.69 | 0.66 | 0.84 | 0.42 | 0.69 | 0.74 | 0.71 |
| | $UNet_{RDM+T}$+CAS+$L_{L2H}$ | 0.65 | 0.71 | 0.68 | 0.83 | 0.40 | 0.72 | 0.73 | 0.73 |
| | Proposed | 0.68 | 0.74 | 0.71 | 0.86 | 0.46 | 0.74 | 0.76 | 0.75 |

Table 6 Quantitative comparisons in Delaware (DE)

| State | Method | mIoU | FWIoU | Kappa | OA | Class | | | |
|---|---|---|---|---|---|---|---|---|---|
| | | | | | | I.S. | L.V. | T.C. | W. |
| DE | DeepLabv3+ | 0.45 | 0.63 | 0.59 | 0.74 | 0.21 | 0.61 | 0.72 | 0.24 |
| | HRNet | 0.47 | 0.58 | 0.52 | 0.70 | 0.22 | 0.53 | 0.66 | 0.48 |
| | L2HNet (RP+CAS+$L_{L2H}$) | 0.61 | 0.68 | 0.63 | 0.74 | 0.42 | 0.62 | 0.69 | 0.70 |
| | RP+ANLC | 0.64 | 0.74 | 0.69 | 0.77 | 0.44 | 0.65 | 0.75 | 0.73 |
| | $UNet_{RDM+T}$+CAS+$L_{L2H}$ | 0.66 | 0.74 | 0.70 | 0.79 | 0.41 | 0.69 | 0.78 | 0.75 |
| | Proposed | 0.69 | 0.76 | 0.73 | 0.83 | 0.44 | 0.70 | 0.80 | 0.79 |

Table 7 Quantitative comparisons in West Virginia (WV)

| State | Method | mIoU | FWIoU | Kappa | OA | Class | | | |
|---|---|---|---|---|---|---|---|---|---|
| | | | | | | I.S. | L.V. | T.C. | W. |
| WV | DeepLabv3+ | 0.53 | 0.66 | 0.62 | 0.75 | 0.23 | 0.61 | 0.77 | 0.51 |

| | | | | | | | | |
|---|---|---|---|---|---|---|---|---|
| HRNet | | 0.52 | 0.63 | 0.59 | 0.74 | 0.26 | 0.54 | 0.76 | 0.54 |
| L2HNet (RP+CAS +$L_{L2H}$) | | 0.60 | 0.71 | 0.70 | 0.77 | 0.39 | 0.65 | 0.76 | 0.61 |
| RP+ANLC | | 0.62 | 0.76 | 0.73 | 0.79 | 0.40 | 0.67 | 0.78 | 0.64 |
| $UNet_{RDM+T}$+CAS+$L_{L2H}$ | | 0.64 | 0.75 | 0.72 | 0.80 | 0.41 | 0.71 | 0.78 | 0.66 |
| Proposed | | 0.66 | 0.78 | 0.74 | 0.82 | 0.45 | 0.72 | 0.81 | 0.67 |

Table 8 Quantitative comparisons using average accuracy scores across the six states.

| Method | mIoU | FWIoU | Kappa | OA | Class | | | |
|---|---|---|---|---|---|---|---|---|
| | | | | | I.S. | L.V. | T.C. | W. |
| DeepLabv3+ | 0.51 | 0.62 | 0.60 | 0.72 | 0.25 | 0.59 | 0.73 | 0.48 |
| HRNet | 0.52 | 0.62 | 0.59 | 0.74 | 0.25 | 0.58 | 0.69 | 0.56 |
| L2hNet (RP+CAS +$L_{L2H}$) | 0.61 | 0.68 | 0.67 | 0.79 | 0.39 | 0.65 | 0.75 | 0.65 |
| RP+ANLC | 0.64 | 0.73 | 0.70 | 0.81 | 0.41 | 0.68 | 0.78 | 0.68 |
| $UNet_{RDM+T}$+CAS+$L_{L2H}$ | 0.65 | 0.73 | 0.71 | 0.82 | 0.41 | 0.70 | 0.79 | 0.69 |
| Proposed | 0.68 | 0.76 | 0.73 | 0.85 | 0.47 | 0.73 | 0.81 | 0.70 |

## 5.3 Ablation Study

In order to validate the effectiveness of each component of the proposed Transformer-based weakly supervised method, we conducted an ablation study.

We present the quantitative results of the following combinations of different modules: (1) U-Net-like Transformer + CE (Cross-Entropy) loss, (2) RDM-based U-Net-like Transformer + CE loss, (3) U-Net-like Transformer + ANLC, (4) RDM-based U-Net-like Transformer + ANLC, which are abbreviated as $UNet_T + L_{CE}$, $UNet_{RDM+T} + L_{CE}$, $UNet_T + ANLC$, and $UNet_{RDM+T} + ANLC$, respectively. According to the results presented in Figure 7 and Table 8, $UNet_T + L_{CE}$ exhibited the poorest visual results and achieved the lowest scores in mIoU, FWIoU, Kappa, and OA. Conversely, $UNet_{RDM+T} + L_{CE}$ achieved better results compared to $UNet_T + L_{CE}$, indicating the positive influence of the RMD-based skip connection. Furthermore, $UNet_{RDM+T} + ANLC$ demonstrated higher accuracy compared to $UNet_T + ANLC$, providing additional evidence to support this observation. By comparing the accuracy results of $UNet_T + L_{CE}$ with $UNet_T + ANLC$, as well as $UNet_{RDM+T} + L_{CE}$ with $UNet_{RDM+T} + ANLC$, it is evident that ANLC significantly enhances the classification outcomes and plays a crucial role in cross-resolution tasks.

Based on the visualized results, it can be observed that the prediction results of $UNet_T + L_{CE}$, as illustrated in Figure 7(f), tend to resemble coarse labels when training the model with the commonly used CE loss. These predictions mostly capture the overall outlines of different categories but fail to accurately represent the internal details. As illustrated in Figure 7(g), the mapping results of $UNet_{RDM+T} + L_{CE}$ exhibit more details compared to those obtained using $UNet_T + L_{CE}$. This validates the positive effect of the integrated RDM which enables the network to capture low-level information. However, it is regrettable that the boundary predictions for roads and buildings are not accurate. On the contrary, as depicted in Figures 7(h) and 7(i), when we integrate the proposed ANLC, the classification results exhibit more details, enabling the accurate prediction of road contours and yielding smoother and more intricate boundary predictions with fewer errors. This is attributed to the effectiveness of ANLC in removing noise from outdated land cover products. In summary, the ablation study demonstrates the effectiveness of the proposed RDM-based U-Net Transformer and ANLC in cross-resolution land cover classification tasks.

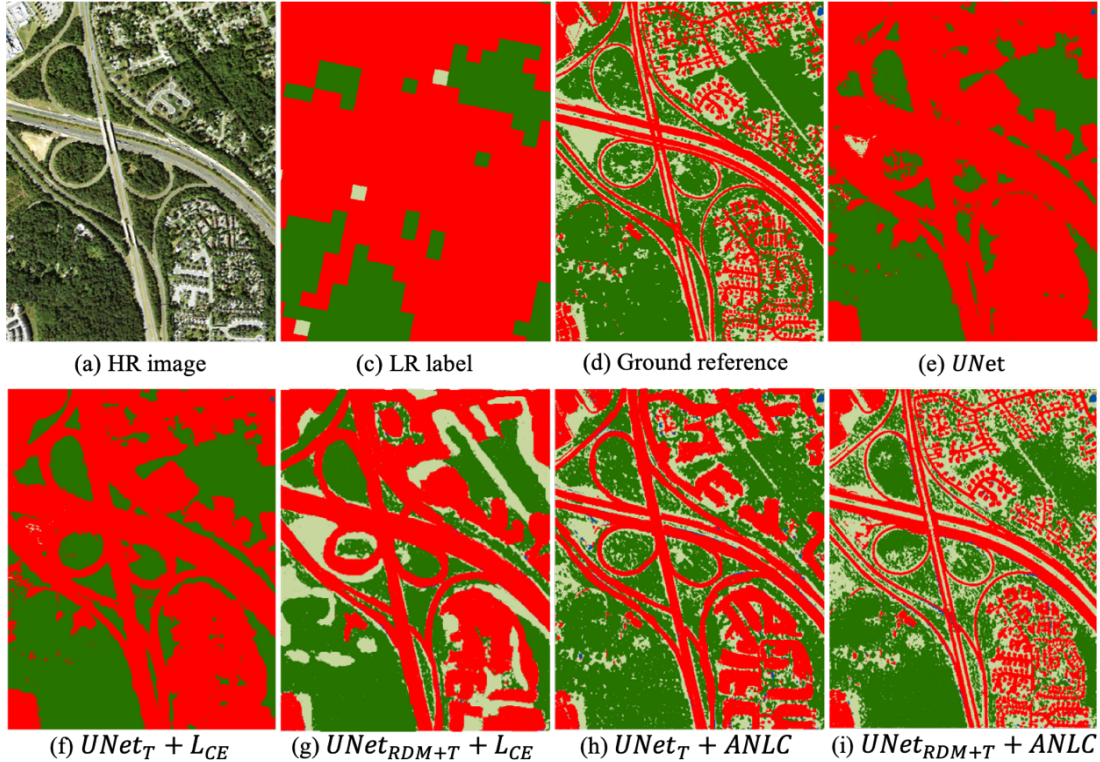

Figure 7 Visual results of the ablation study

Table 8 Ablation results for the proposed framework with different optimization settings on the Chesapeake Bay watershed datasets

| Metric | Optimization setting | States | | | | | |
|---|---|---|---|---|---|---|---|
| | | ML | PA | VA | NY | DE | WV |
| mIoU | $UNet_T + L_{CE}$ | 0.43 | 0.44 | 0.47 | 0.44 | 0.49 | 0.46 |
| | $UNet_{RDM+T} + L_{CE}$ | 0.53 | 0.56 | 0.54 | 0.55 | 0.50 | 0.55 |
| | $UNet_T + ANLC$ | 0.66 | 0.64 | 0.62 | 0.65 | 0.67 | 0.63 |
| | $UNet_{RDM+T} + ANLC$ | 0.70 | 0.66 | 0.68 | 0.68 | 0.69 | 0.66 |
| FWIoU | $UNet_T + L_{CE}$ | 0.44 | 0.44 | 0.57 | 0.54 | 0.55 | 0.57 |
| | $UNet_{RDM+T} + L_{CE}$ | 0.58 | 0.58 | 0.63 | 0.60 | 0.55 | 0.59 |
| | $UNet_T + ANLC$ | 0.72 | 0.71 | 0.72 | 0.72 | 0.73 | 0.77 |
| | $UNet_{RDM+T} + ANLC$ | 0.76 | 0.74 | 0.76 | 0.74 | 0.76 | 0.78 |
| Kappa | $UNet_T + L_{CE}$ | 0.37 | 0.35 | 0.54 | 0.53 | 0.53 | 0.55 |
| | $UNet_{RDM+T} + L_{CE}$ | 0.58 | 0.60 | 0.63 | 0.61 | 0.54 | 0.58 |
| | $UNet_T + ANLC$ | 0.73 | 0.72 | 0.69 | 0.67 | 0.70 | 0.74 |
| | $UNet_{RDM+T} + ANLC$ | 0.75 | 0.73 | 0.72 | 0.71 | 0.73 | 0.74 |
| OA | $UNet_T + L_{CE}$ | 0.52 | 0.57 | 0.69 | 0.67 | 0.68 | 0.69 |
| | $UNet_{RDM+T} + L_{CE}$ | 0.72 | 0.73 | 0.75 | 0.73 | 0.69 | 0.71 |

| | | | | | | |
|---|---|---|---|---|---|---|
| $UNet_T + ANLC$ | 0.82 | 0.81 | 0.83 | 0.85 | 0.80 | 0.80 |
| $UNet_{RDM+T} + ANLC$ | 0.88 | 0.84 | 0.87 | 0.86 | 0.83 | 0.82 |

# 6 Conclusion

With the increasing demand for more detailed and more extensive land cover products, it is important to use readily available outdated LR land cover products to accomplish large-scale HR land cover classification. This can greatly reduce the need for manual labor in updating land cover classification. However, due to the mismatch issue in terms of resolution and time, this task is highly challenging. This study proposes a Transformer-based weakly supervised method for cross-resolution land cover classification using outdated products. First, to capture long-range dependencies and mitigate the spatial information loss caused by down-sampling, we propose an RDM-based U-Net Transformer with dynamic sparse attention for effective feature representation. The RDM-based U-Net Transformer extracts both global information and spatial details which help deal with the cross-resolution tasks. Second, to relieve the impact of noise in the outdated products, an anti-noise loss calculation module is proposed based on OT. Anti-noise loss calculation detects the reliable correspondences between predictions and outdated products, and divides the coarse labels into CA and VA. Then, the weakly supervised loss with weights is constructed.

The experiments were conducted using a large amount of data covering six states of the Chesapeake Bay watershed in the United States. To train the model, we used HR aerial images with 1 m resolution from 2017 as the datasets, and utilized the LR land cover products with 30 m resolution from 2013 as the supervision source. Under this cross-resolution and cross-time phase settings, the experimental results were obtained. The experimental results show that the proposed Transformer-based weakly supervised method significantly improves the classification accuracy compared to the fully supervised methods. Moreover, the Transformer-based weakly supervised method outperforms the existing methods designed for the cross-resolution land cover classification task. In addition, the ablation study validates the great influences of the RDM-based U-Net Transformer and ANLC.

In terms of the limitations of this study, the unsupervised loss for VA was directly employed from [7]. There is still room for improvement of the unsupervised loss. In future, more reliable unsupervised losses for VA will be studied.